\documentclass[conference]{IEEEtran}
\IEEEoverridecommandlockouts
\usepackage{cite}
\usepackage{amsmath,amssymb,amsfonts}
\usepackage{algorithmic}
\usepackage{graphicx}
\usepackage{textcomp}
\usepackage{xcolor}
\def\BibTeX{{\rm B\kern-.05em{\sc i\kern-.025em b}\kern-.08em
    T\kern-.1667em\lower.7ex\hbox{E}\kern-.125emX}}
\begin{document}

\title{Historical Knowledge Graphs for Global Maritime Estimated Time of Arrival\\
\thanks{This work was co-financed by the European Union—NextGenerationEU, through the Research and Innovation Foundation with grant numbers CODEVELOP-GT/0322/0096 (ADAPTATION) and STRATEGIC INFRASTRUCTURES/1222/0113 (MDigi-I), and the EU H2020 Research and Innovation Programmes under Grant Agreements No. 857586 (CMMI-MaRITeC-X).}
}

\author{\IEEEauthorblockN{Neofytos Dimitriou}
\IEEEauthorblockA{\textit{Maritime Digitalization Centre} \\
\textit{Cyprus Marine and Maritime Institute}\\
Larnaca 6300, Cyprus\\
0000-0002-2328-6502}
}

\maketitle

\begin{abstract}
Accurate vessel estimated-time-of-arrival forecasts are critical for port operations and decarbonization, yet global-scale travel-time prediction remains difficult without costly contextual data. Herein, I present a methodology for constructing a historical maritime knowledge graph using only Automatic Identification System (AIS) data. First, segmented trajectories are extracted from noisy AIS data using a Gaussian-mixture-model-based preprocessing pipeline. The graph is then constructed by iteratively processing the trajectories and storing speed distributions stratified by vessel type, time of travel, and direction of travel; the resulting global graph comprises $5,433$ geohash-3 nodes and $12,334$ edges. The graph can be queried to retrieve travel-time predictions between any two location via a hierarchical, priority‑based system that uses historical statistics with principled fallback. On a temporally held-out test set, median RMSE is $22.75$ min (segment-level) and $30.90$ min (trajectory-level), with $69.1\%$ of trajectories within $20\%$ of actual arrival time. On a second external test set, median RMSE is $27.36$ min (segment-level) and $37.46$ min (trajectory-level), with $62.1\%$ of trajectories within $20\%$. These results corroborate the promise of our method, enabling global travel-time prediction and providing a strong foundation for just-in-time arrival planning and emissions reduction.
\end{abstract}


\section{Introduction}
\label{sec:intro}
Reliable maritime ETA prediction underpins port efficiency, just-in-time arrival strategies, and emissions reduction. Achieving accuracy at global scale is challenging: a fully general solution would integrate weather, currents, and port conditions, but such inputs are often unavailable, inconsistent, or prohibitively expensive worldwide. AIS is the only truly ubiquitous signal; yet even historical AIS data is not free at scale (commercial providers charge substantial fees), and community real‑time feeds (e.g. AISStream\footnote{https://aisstream.io/}) do not cover open-ocean navigation. Shipping companies do hold proprietary AIS archives for their fleets and nearby traffic, but these data are rarely shared or are shared under terms that preclude publication. 

\begin{figure*}[b!th]
\includegraphics[width=1.0\linewidth]{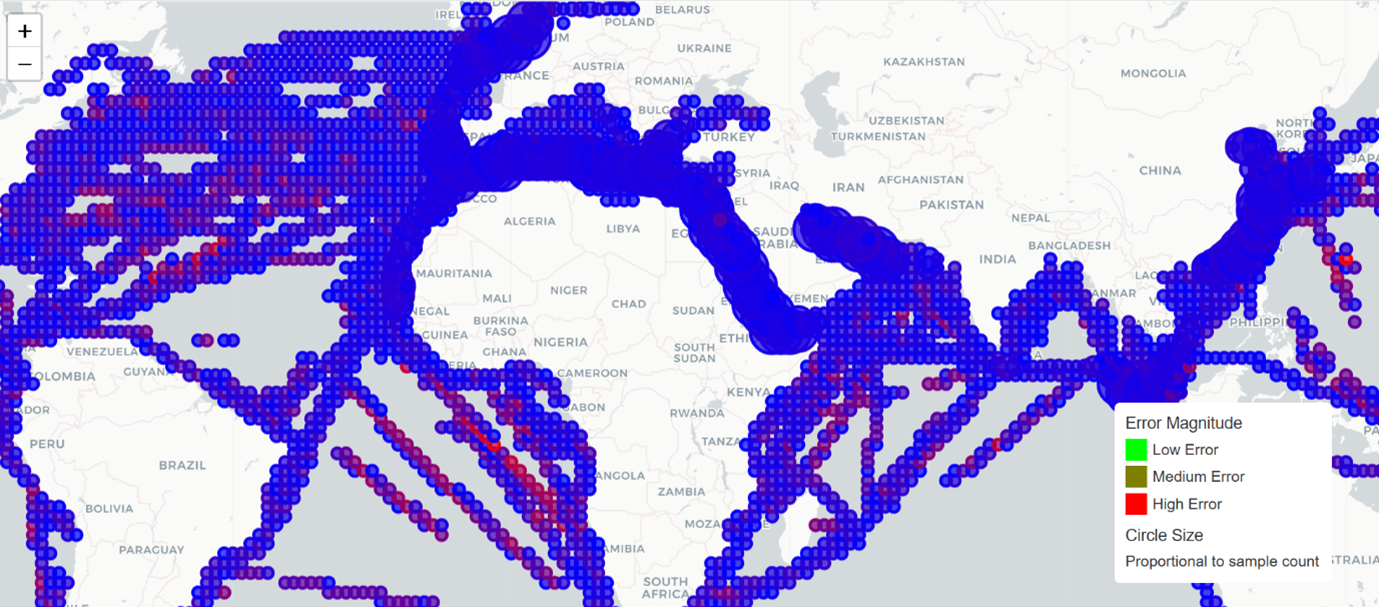}
\caption{Spatial visualization of the average per-node errors of the test set trajectories.}
\label{fig:res}
\end{figure*}

This work leverages AIS-only data, distilling historical movement patterns into an open-access, graph-based knowledge base that can be iteratively enriched with information from closed-access sources. A dataset-adaptive preprocessing pipeline based on Gaussian Mixture Models is introduced that automatically identifies ships with sufficient travelling data. By processing their trajectories, a historical maritime knowledge graph is constructed; context‑specific speed information is accumulated based on vessel type, temporal context, and direction of travel. The approach scales globally: from five months of AIS data (March-July 2023), $2.59$ B dynamic messages from $140,867$ vessels are processed to build a global graph with $5,433$ geohash-3 nodes and $12,334$ edges.

The graph supports time-travel prediction via a hierarchical, priority-based estimator that favours the most specific historical statistics and degrades gracefully when data are sparse. 
Our contributions are: (1) A dataset-adaptive AIS preprocessing and segmentation pipeline that extracts high-quality trajectories from consistent ship transmitters without manual thresholds. (2) A global-scale maritime knowledge graph that captures context-specific speed distributions and which can be released publicly as a reusable asset. (3) A hierarchical estimation scheme that provides robust coverage with principled fallback. (4) Extensive evaluation and global-scale validations demonstrating strong accuracy with AIS-only inputs (see Figure~\ref{fig:res}). The released knowledge base can be continually augmented with additional trajectories, and be used as the foundation for training more complex models (e.g.\ graph neural networks, see the work by Derrow-Pinion et al.~\cite{DerrSheWongLang+2021}), enabling both industrial use and open research while avoiding redistribution of private raw data.

\section{Related Work}
A recent survey organizes vessel ETA research into two broad families: (i) non-trajectory-based prediction and (ii) trajectory-based prediction via path finding and travel-time estimation~\cite{JianLiuPengXu+2025}.

\subsection{Non-trajectory-based approaches}
These methods forecast arrival time directly from tabular features (AIS dynamic data, vessel and route/port static data, and weather data), typically per port or route~\cite{JianLiuPengXu+2025}. For example, deep learning models including recurrent and convolutional neural networks have been previously trained using AIS, vessel, and weather data to predict travel-time of bulk carriers to a destination port from different routes~\cite{MekkBenaBerr2023}. Evmides et al.~\cite{EvmiAslaRameMich+2024} performed feature selection and hyperparameter tuning of several machine learning models, providing a comparative analysis between their performance in predicting time of arrival to the port of Limassol. Although competitive locally, these models have not been demonstrated to generalize globally.

\subsection{Trajectory-based approaches}
Trajectory-based methods couple path-finding with segment travel-time estimation. Alessandrini et al.~\cite{AlesMazzVesp2019} combine AIS and Long Range Identification and Tracking (LRIT) data to build raster maps (traffic density, preferred direction, land mask) and apply a least-cost path search; travel-time is computed from historical speed statistics along the inferred path~\cite{AlesMazzVesp2019}, and the method was validated at the port of Trieste. Park et al.\ approach path‑finding through reinforcement learning and over an AIS‑derived grid, while speed estimation is done via MCMC sampling from direction‑conditioned speed-over-ground distributions. ETA is then the distance integral over the expected speed along the learned path~\cite{ParkSimBae2021}. Experiments use AIS data around the Port of Busan, South Korea, showing higher path-finding success and improved trajectory similarity versus density-based baselines~\cite{ParkSimBae2021}. Zhai et al.\ propose probability-density scanning to reconstruct generic trajectories between origin-destination pairs, then compute ETA via great-circle distance and a constant-speed approximation~\cite{DeqiXiujXiaoHaiy+2022}. The proposed method was evaluated on long-haul routes between Singapore and Australian ports (Adelaide, Brisbane, Perth). Prior trajectory-based work is largely port/region-specific, underscoring the gap to site-agnostic global travel-time prediction.

I follow the trajectory-based paradigm but avoid local modelling and additional data sources. Instead, a global geohash-3 graph is constructed that stores context-specific historical speed distributions (by vessel type, time, and direction), and query it with a hierarchical fallback for segment times, enabling AIS-only travel-time estimation at global scale.
\section{Proposed Method}
\begin{figure*}[tb!h]
\includegraphics[width=1.0\linewidth]{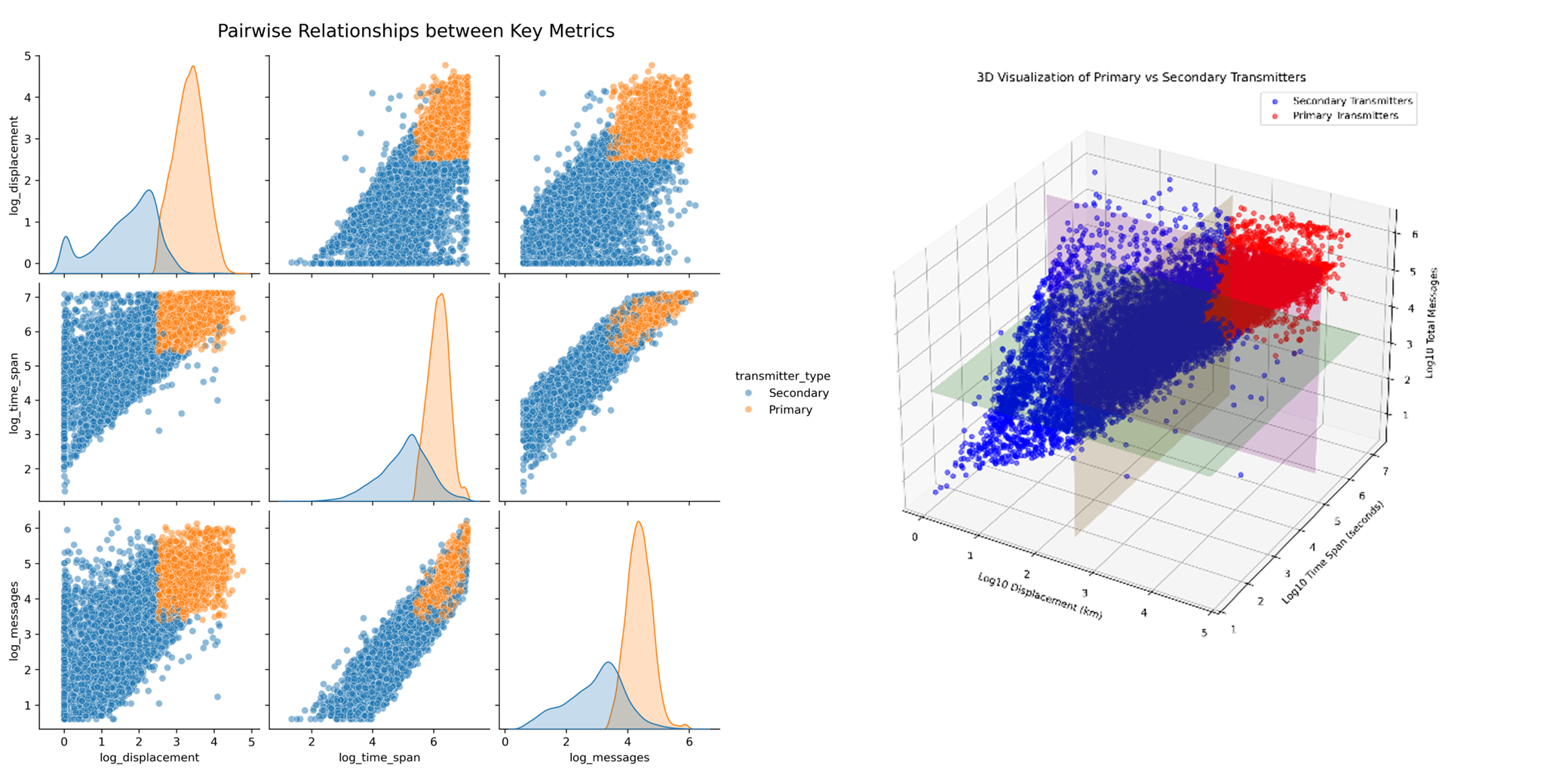}
\caption{(Left) Pairwise relationships of  displacement, time\_span and num\_messages between the two dynamically set clusters. (Right) 3D visualization of individual ships based on the three metrics, with colour indicating whether they were clustered as primary or secondary transmitters; the latter is removed from the training data.}
\label{fig:preprocessing}
\end{figure*}
\subsection{Dataset and Preprocessing}
I use five months of Automatic Identification System (AIS) data (March-July 2023): $2.59$ billion dynamic messages from $140,867$ unique vessels, provided by Tototheo Ltd under the ADAPTATION project (CODEVELOP-GT/0322/0096; European Union's NextGenerationEU). Most messages are not suitable for trajectory analysis because they originate from non-target vessels transiting near the main fleet. Conversely, restricting to the main fleet via manually tuned thresholds discards substantial amounts of relevant data. To address this, I adopt a dataset-adaptive pipeline that (i) segments raw trajectories into eligible sub-trajectories and (ii) selects consistent, high-quality transmitters (beyond just the main fleet) using unsupervised modelling.

For objective evaluation, we apply a chronological split: the last week of each month is held out for testing; the remainder is used to fit preprocessing models and build the knowledge graph. Modelling hyper-parameters (including mixture models below) are fit only on training data and then applied to the held‑out weeks and future data.

Per vessel, I first partition the time-ordered stream of dynamic AIS messages into sub-trajectories by starting a new sub-trajectory whenever the gap between consecutive messages exceeds 90 minutes. I then filter messages to keep only those with reported speed over ground in between 3 and 50 knots. A sub-trajectory is deemed eligible iff, after this speed filtering, it satisfies all of the following: (i) it contains $\ge 10$ dynamic messages; (ii) its travel time is $\ge 30$ minutes; and (iii) its displacement is $\ge 100$ km. Finally, each eligible sub-trajectory is summarized by displacement, time span, and message count - features later used for transmitter selection.

Fixed global thresholds discard too much data and introduce geographic/ship type biases depending on the available data. Instead, I fit Gaussian Mixture Models (GMMs) on sub-trajectories summaries from the training set to discover natural strata of reporting behaviour and automatically separate primary transmitters (consistent, high‑quality reporters) from secondary ones. The learned decision boundaries are then applied to the held-out weeks and any future data. After selection, we retain $38,181$ high‑quality vessels (as opposed to less than $5,000$ with manually set thresholds) contributing $2,083,955$ sub-trajectories for analysis (see Fig.~\ref{fig:preprocessing}).

\subsection{Historical Maritime Knowledge Graph Construction}
I model the global maritime space as a spatiotemporal knowledge graph KG = (V, E, T) where nodes represent spatial regions, edges capture bidirectional transitions, and temporal dimensions store historical speed distributions.

Nodes (V): Each node corresponds to a geohash cell at precision 3 (approximately 156km $\times$ 156km at the equator). This discretization treats each cell as a segment within a coarse maritime network, analogous to road segments~\cite{DerrSheWongLang+2021} but adapted for open-ocean navigation. Edges (E): Directed edges connect geohash cells that are visited consecutively by vessels (within a sub-trajectory), capturing observed spatial transitions. In this work, edge structure is used primarily to facilitate retrieval of relevant information; in future work, distinct information and features may be attached to nodes and edges. Temporal Dimension (T): Each node maintains empirical speed distributions stratified by three key factors: (i) Ship type (cargo, tanker, other), (ii) Temporal context (hour-of-day, day-of-week, month-of-year), and (iii) Direction of travel (eight compass directions).

This work's core modeling assumption posits that vessels sharing the same contextual tuple (ship-type, temporal-context, heading) within a given geohash cell exhibit similar speed profiles. This enables context-specific speed priors for travel-time calculations rather than relying on e.g.~global averages.

Sub-trajectories are processed in time order. Each message is mapped to a fixed-precision geohash cell (graph nodes). While consecutive messages remain in the same cell, I accumulate speed statistics. Upon a transition to a new cell, we flush the accumulator: compute (sum, sum of squares, min, max, count) for the run just completed and add these to (i) the source node (the cell just left) and (ii) the directed edge from source to destination. Nodes and edges are created lazily on first observation. Segments that never leave their initial cell are ignored. Using this procedure, I processed the training sub-trajectories, yielding a graph with $5,433$ nodes and $12,334$ directed edges.

\subsection{Priority-Based Speed Estimation}
\label{sec:priority}
Travel-time predictions leverage hierarchical priority levels for speed estimation, progressively relaxing specificity when historical data is sparse:
\begin{itemize}
    \item Group 1 (Highest Specificity): Exact matching of direction, ship type, and one of the temporal stratification, i.e.\ exact hour-of-day, day-of-week, or month.
    \item Group 2: Exact matching of direction and ship type, and aggregating statistics across time. If not available, aggregating statistics across both time and ship type.
    \item Group 3: Aggregating statistics across directions while maintaining ship type and one of the temporal stratification.
    \item Group 4 (Lowest Specificity): Exact matching of either direction, ship type, or temporal features, while aggregating over the remaining.
    \item Fallback: Standard vessel class speeds when no historical data exists.
\end{itemize}

The system queries starting from highest specificity, progressively broadening criteria until finding suitable historical data. Estimates are classified as ``reliable'' when supported by $\ge 8$ historical observations, balancing statistical robustness with data availability. This priority system ensures optimal use of available information while gracefully degrading in data-sparse regions.

\section{Evaluation}
\subsection{Testing sets}
As previously mentioned, a temporal train-test splitting is adopted, reserving the final week of each month for testing while using preceding weeks for the graph creation. The temporally held-out test set comprises $39,610$ sub-trajectories segmented into $76,889$ individual transitions between geohash cells. Furthermore, a second test set is used which contains one month of AIS data from February 2025 ($25,480$ sub-trajectories, $59,743$ individual transitions between geohash cells). 



\subsection{Evaluation}
On the temporally held-out test set, historical-only estimation (Groups 1–4; Sec. \ref{sec:priority}) covered $99.6\%$ of segments and $99.7\%$ of sub-trajectories. For those segments, RMSE (median/mean) is $22.75$/$32.26$ min and MAE (median/mean) is $29.63$/$20.55$ min; $68.06\%$ of estimates were within $20\%$ of ground truth, $37.57\%$ within $10\%$, and $19.21\%$ within $5\%$. At the trajectory level, across all trajectories, we report an RMSE (median/mean) of $30.90$/$56.42$ min; $19.32\%$ of estimates were within $5\%$ of ground truth, $37.85\%$ within $10\%$, and $69.13\%$ within $20\%$. For long trajectories ($>150$ km), RMSE (median/mean) is $46.86$/$79.45$ min; $19.72\%$ were within $5\%$, $38.43\%$ within $10\%$, and $70.32\%$ within $20\%$.

On the temporally out-of-distribution test set, historical-only estimation covered $94.4\%$ of segments and $97.0\%$ of sub-trajectories. For those segments, RMSE (median/mean) is $27.36$/$37.23$ min and MAE (median/mean) is $24.50$/$33.76$ min. In total, $60.65\%$ of estimates were within $20\%$ of ground truth, $32.34\%$ within $10\%$, and $16.57\%$ within $5\%$. At the trajectory level, we report an RMSE (median/mean) of $37.46$/$76.79$ min; $62.13\%$ of estimates were within $20\%$, $33.08\%$ within $10\%$, and $17.09\%$ within $5\%$.

\section{Conclusions}
This work demonstrates that accurate global-scale maritime travel-time predictions are achievable to a large extent using exclusively historical AIS data, without requiring weather information, port-specific models, or real-time updates. Our historical maritime knowledge graph successfully captures complex spatiotemporal patterns of global shipping, enabling context-aware speed estimation that adapts to vessel type, temporal factors, and geographic location. 

Future work will explore integration of weather data and port congestion information where available, both historical and real-time, and investigate data-driven modelling (e.g. graph neural networks) for learning complex spatial dependencies. The knowledge graph framework provides a solid foundation for these enhancements while maintaining the core capability of global predictions under data constraints.
\bibliographystyle{IEEEtran}
\bibliography{refs}

@article{JianLiuPengXu+2025,
author = {Shuo Jiang and Lei Liu and Peng Peng and Mengqiao Xu and Ran Yan},
title = {Prediction of vessel arrival time to port: a review of current studies},
journal = {Maritime Policy \& Management},
volume = {0},
number = {0},
pages = {1--26},
year = {2025},
publisher = {Routledge},
doi = {10.1080/03088839.2025.2488376}
}

@Article{MekkBenaBerr2023,
author={El Mekkaoui, Sara
and Benabbou, Loubna
and Berrado, Abdelaziz},
title={Deep learning models for vessel's ETA prediction: bulk ports perspective},
journal={Flexible Services and Manufacturing Journal},
year={2023},
month={Mar},
day={01},
volume={35},
number={1},
pages={5-28},
issn={1936-6590},
doi={10.1007/s10696-022-09471-w}
}

@Article{EvmiAslaRameMich+2024,
AUTHOR = {Evmides, Nicos and Aslam, Sheraz and Ramez, Tzioyntmprian T. and Michaelides, Michalis P. and Herodotou, Herodotos},
TITLE = {Enhancing Prediction Accuracy of Vessel Arrival Times Using Machine Learning},
JOURNAL = {Journal of Marine Science and Engineering},
VOLUME = {12},
YEAR = {2024},
NUMBER = {8},
ARTICLE-NUMBER = {1362},
URL = {https://www.mdpi.com/2077-1312/12/8/1362},
ISSN = {2077-1312},
DOI = {10.3390/jmse12081362}
}

@ARTICLE{AlesMazzVesp2019,
  author={Alessandrini, Alfredo and Mazzarella, Fabio and Vespe, Michele},
  journal={IEEE Transactions on Intelligent Transportation Systems}, 
  title={Estimated Time of Arrival Using Historical Vessel Tracking Data}, 
  year={2019},
  volume={20},
  number={1},
  pages={7-15},
  doi={10.1109/TITS.2017.2789279}
}

@article{ParkSimBae2021,
title = {Vessel estimated time of arrival prediction system based on a path-finding algorithm},
journal = {Maritime Transport Research},
volume = {2},
pages = {100012},
year = {2021},
issn = {2666-822X},
doi = {https://doi.org/10.1016/j.martra.2021.100012},
url = {https://www.sciencedirect.com/science/article/pii/S2666822X21000046},
author = {Kikun Park and Sunghyun Sim and Hyerim Bae},
}

@article{DeqiXiujXiaoHaiy+2022,
  publtype={informal},
  author={Deqing Zhai and Xiuju Fu and Xiaofeng Yin and Haiyan Xu and Wanbing Zhang and Ning Li},
  title={Constructing Trajectory and Predicting Estimated Time of Arrival for Long Distance Travelling Vessels: A Probability Density-based Scanning Approach},
  year={2022},
  cdate={1640995200000},
  journal={CoRR},
  volume={abs/2205.07945},
  url={https://doi.org/10.48550/arXiv.2205.07945}
}

@article{DerrSheWongLang+2021,
  title={ETA Prediction with Graph Neural Networks in Google Maps},
  author={Austin Derrow-Pinion and Jennifer She and David Wong and Oliver Lange and Todd Hester and Luis Perez and Marc Nunkesser and Seongjae Lee and Xueying Guo and Brett Wiltshire and Peter W. Battaglia and Vishal Gupta and Ang Li and Zhongwen Xu and Alvaro Sanchez-Gonzalez and Yujia Li and Petar Velivckovi'c},
  journal={Proceedings of the 30th ACM International Conference on Information \& Knowledge Management},
  year={2021},
  url={https://api.semanticscholar.org/CorpusID:237303762}
}
\end{document}